\documentclass[a4paper]{article}

\usepackage{INTERSPEECH_v2}

\usepackage{amsmath,graphicx}
\usepackage{dblfloatfix} 
\usepackage{multirow}
\usepackage{tabularx}
\usepackage{array,booktabs}
\usepackage{threeparttable}
\usepackage{subfig}
\usepackage{url}
\usepackage{tipa}
\usepackage[svgnames]{xcolor}
\usepackage{mathrsfs} 
\usepackage{mathtools}
\usepackage{tensor}
\usepackage{hyperref}
\usepackage{textcomp}
\usepackage{soul}

\ninept

\newcolumntype{L}[1]{>{\raggedright\let\newline\\\arraybackslash\hspace{0pt}}m{#1}}
\newcolumntype{C}[1]{>{\centering\let\newline\\\arraybackslash\hspace{0pt}}m{#1}}
\newcolumntype{R}[1]{>{\raggedleft\let\newline\\\arraybackslash\hspace{0pt}}m{#1}}

\newcommand{\argmax}[1]{\underset{#1}{\operatorname{arg}\,\operatorname{max}}\;}

\title{Machine Assisted Analysis of Vowel Length Contrasts in Wolof}

\name{Elodie Gauthier$^1$, Laurent Besacier$^1$, Sylvie Voisin$^2$}
\address{
  $^1$Laboratoire d'Informatique de Grenoble (LIG), Univ. Grenoble Alpes, Grenoble, France\\
  $^2$Laboratoire Dynamique Du Langage (DDL), CNRS - Universit\'e Aix Marseille, France}
\email{elodie.gauthier@imag.fr, laurent.besacier@imag.fr, sylvie.voisin@cnrs.fr}

\begin{document}
%
\maketitle

\begin{abstract}

Growing digital archives and improving algorithms for automatic analysis of text and speech create new research opportunities for fundamental research in phonetics. 
Such empirical approaches allow statistical evaluation of a much larger set of hypothesis about phonetic variation and its conditioning factors (among them geographical / dialectal variants).
This paper illustrates this vision and proposes to challenge automatic methods for the analysis of a not easily observable phenomenon: vowel length contrast.  We focus on Wolof, an under-resourced language from Sub-Saharan Africa. In particular, we  propose multiple features to make a fine evaluation of the degree of length contrast under different factors such as: read \textit{vs} semi-spontaneous speech ; standard  \textit{vs}  dialectal Wolof. 
Our measures made fully automatically on more than 20k vowel tokens show that our proposed features can highlight different degrees of contrast for each vowel considered. We notably show that contrast is weaker in semi-spontaneous speech and in a non standard semi-spontaneous dialect.

\end{abstract}

\vspace{-0.2cm}
\section{Introduction}


Growing digital archives and improving algorithms for automatic analysis of text and speech create new research opportunities for fundamental research in linguistics and phonetics. 
This vision is shared by  \cite{lieberman2016} where audiobooks (large amount of recordings in many languages and dialects, distributed in a natural way across a wide variety of speakers) are used for corpus-based phonetics. In their work, authors claim that - for the phonetic events observed - \textit{``the data used from audiobooks offers more tokens than have been examined in the entire 50-year history of sociolinguistic study of Spanish''}. 
In a similar trend, we have recently shown the value of stochastic and neural acoustic models for analyzing, at a relatively large scale, vowel length contrast in two under-resourced african languages \cite{interspeech-gauthier2016}. 
Such empirical approaches allow statistical evaluation of a much larger set of hypothesis about phonetic variation and its conditioning factors (among them geographical / dialectal variants).
This paper illustrates this vision and proposes a detailed analysis of vowel length constrast in Wolof under different factors such as: read \textit{vs} semi-spontaneous speech ; standard (Dakar) Wolof 
\textit{vs}  dialectal (Faana-Faana) Wolof. 

	\textbf{Paper contributions. }  The first contribution of this paper is a large scale analysis of vowel length contrast on Wolof read speech. Multiple features are proposed to judge the degree of bimodality in the distribution (of durations) for a given vowel. Our measures made on 14k vowel tokens show different degrees of contrast according to the vowel considered. We also show, in a second contribution, that in the case of read speech, the need of manual transcriptions can be relaxed since the use of automatic speech recognition (ASR) can lead to very similar measurements and to the same conclusions.  Our third contribution is an application of our machine-assisted methodology to study vowel length contrast in more spontaneous speech for Wolof and for one of its dialectal variant (Faana-Faana). For reproductible research, a Wolof ASR VM and the data of this study are also made available online\footnote{see \url{https://github.com/besacier/ALFFA_PUBLIC/blob/master/ASR/WOLOF/WOLOF-VM/} and \url{https://github.com/besacier/ALFFA_PUBLIC/blob/master/ASR/WOLOF/INTERSPEECH_2017}}.
\\

	\vspace{-0.2cm}
	\textbf{Languages studied. }
Wolof is the vehicular language of Senegambia (Senegal and Gambia), also spoken in Mauritania. This paper focuses on senegalese Wolof. 
We will use the term ``standard'' to refer to Wolof spoken in Dakar by native speakers of the language and ``urban'' for Wolof spoken by non-native speakers.
In Senegal, there are also dialectal variants but mutual understanding exists between people living in the different areas. Linguists observe some phonetic or morpho-phonological variations, focusing on vocalism, on some forms of verbal inflection \cite{robert:hal-00600630} and also on some morphological and syntactical variations \cite{voisin2017inacc}, \cite{voisin2017wolofvar}.

The Faana-Faana dialect studied in this paper is spoken in the region of Kaolack, also named \textit{Wolof of the Saloum}. It is described by Dram\'e \cite{drame2012phonologie} and is
 closer to the Wolof of Gambia. 
This regional variant is not much influenced by other Wolof dialects. However, young people and men often spend part of their lives in Dakar and come back with influences from standard Wolof. Faana-Faana speakers live in a predominant Sereer speaking area which influences their own language, but they are not subject to other major linguistic influences.


In Wolof, the vocalic system is composed of 8 short vowels /i/, /e/, \textipa{/E/}, /a/, \textipa{/@/}, \textipa{/O/}, /o/, /u/; each having a long counterpart (except \textipa{/@/}).
There is no tone in Wolof but phonemes can vary in length \cite{cisse2006problemes}. This means that word sense may differ depending on phoneme duration. For instance, the pronunciation of ``fit'' (\textit{bravery}) and ``fiit'' (\textit{trap}) varies only at the vowel length level, as well as ``wall'' (\textit{to rescue}) 
 and ``waal'' (\textit{to take advantage of}), or ``set'' (\textit{to be clean}) and ``seet'' (\textit{to look for}). Same short and long vowels
exists in the Faana-Faana variant. 
 As can be seen in the examples above, reduplication of the vowel, in the spelling of Wolof, encodes the duration. 
One goal of this paper is to verify if this expected (phonological) contrast is also observed at the phonetic level.
\\
	
	\vspace{-0.2cm}

	\textbf{Paper outline. }
	This paper is organized as following. Section 2 reviews previous works on phonemic contrast analysis. In Section 3, we propose several features to measure degree of (length) contrast for a given unit. 
	In Section 4, we present our multi factor analysis of vowel length contrast in Wolof read and semi-spontaneous speech (Dakar and Faana-Faana).  Finally, Section 5 concludes this work and gives some perspectives.

\section{Related Works}


Vowel duration is a phonetic measure widely used in speech acoustic research. Many factors affect vowel duration such as its location within the vowel space (\cite{lindblom1967vowel}, \cite{lehiste1970suprasegmentals}), position and length of the word \cite{lindblom1981durational}, surrounding context of the vowel (\cite{house1961vowel}, \cite{maddieson1984phonetic}), speech rate (\cite{gay1981mechanisms}, \cite{magen1993effects}) and position of the vowel within the word \cite{myers2005vowel}.
As raised by \cite{adi2016automatic}, main past studies of vowel duration were done through manual annotations. It is consequently a very time-consuming task and only few words were generally analyzed. We believe that use of automatic tools can lead to more objective and reproductible measures, at a larger scale. 
As far as vowel length contrast is concerned, \cite{lee2016acoustic} studied its production and perception in Korean. They found that all Korean speakers of the study produced (length) contrasted vowels but they also concluded that short/long contrast is weaker in spontaneous speech. 
Vowel length contrast was also investigated to better understand language acquisition. \cite{bion2013learning} analyzed 11 hours of Japanese infant-directed speech, using statistical methods, to explore how infants learn to discriminate vowel length contrast existing in Japanese. They discovered that duration distribution for a given vowel is not clearly bi-modal since long vowels may be much less frequent than short vowels.

In Wolof, very few phonetic studies were published, especially on vowel length contrast. 
One exception is the work of \cite{sock1983organisation} who studied a dialectal variant of Gambian Wolof, close to  Faana-Faana analyzed in this paper. 
The author compared 3 minimal pairs, each containing /i/, /a/ and /u/ vowels (read speech) and noticed that length contrast was more important for vowel /a/  than for /i/ and /u/. Moreover, 
less (length) contrast was observed in rapid speech rate compared to normal speech rate.
Finally, in 2006, \cite{cisse2006problemes} pointed out that a large analysis of Wolof phonetics was lacking and to the best of our knowledge this is still the case at present.




		


\section{Measuring Vowel Length Contrast}
\label{features}
It is not trivial to objectively analyze  the degree of bimodality in the distribution of durations for a given vowel. One reason is that - for some vowels - there may be much more short occurences than long ones \cite{sauvageot1965description}. Eye-looking at distributions is a possibility but more objective features are needed if we want a fine evaluation of the degree of contrast across different speech styles and dialects (see \cite{bion2013learning} for Japanese).
This section proposes different criteria (features) to estimate the degree of bimodality for the (duration) distribution of a given vowel. These features are not extracted from true distributions of short and long vowels, but from their normalized gamma approximations\footnote{We preferred \textit{Gamma} distributions to \textit{Gaussian} for their skewness.} - see Figure 1 for the notations used:  (1) ratio $r_1$, (2) ratio $r_2$, (3) area $\mathscr{A}$ between both (short/long) gamma distributions and (4) delta $\Delta$ between modes of both gamma distributions.

We define $d_S(x)$ and $d_L(x)$ as representing respectively the distribution of the short and long units of a vowel (for instance $d_{/i/}(x)$ and $d_{/ii/}(x)$).
In accordance with this definition, $r_1$ is defined by equation \eqref{r1_ratio} and is the ratio between  $d_S(a)$ and $d_L(a)$, when $a$ is the global maximum value of $d_S(x)$. A high value of $r_1$ means a large amount of short tokens compared to long tokens at the maximum peak of $d_S(x)$.
In the same way, $r_2$ defined in equation \eqref{r2_ratio} is the ratio between $d_L(b)$ and $d_S(b)$, when $b$ is the global maximum value of $d_L(x)$. A high value of $r_2$ means a large amount of long tokens compared to short tokens at the maximum peak of $d_L(x)$.
For both ratios, the bigger the value, the stronger the duration contrast is.

\begin{equation}
r_1 = \frac{d_S(a)}{d_L(a)}
\label{r1_ratio}
\end{equation}
where {
$
a = \argmax{x}(d_S(x))
$
.} 

\begin{equation}
r_2 = \frac{d_L(b)}{d_S(b)}
\label{r2_ratio}
\end{equation}
where {
$
b = \argmax{x}(d_L(x))
$
.} \\

$\mathscr{A}$ corresponds to the computed area between both curves when $d_S(x) < d_L(x)$, as shown in equation \eqref{area_equation}. The larger the area, the  stronger the duration contrast should be. We consider that a significant contrast should give an area $\mathscr{A} > 0.40 $.

\begin{equation}
\mathscr{A} = \int_I^\infty \! d_L(x) - d_S(x) \, \mathrm{d}x
\label{area_equation}
\end{equation}

We also compute $\Delta$ which is the difference between both modes of $d_S(x)$ and $d_L(x)$, as represented in equation \eqref{delta_equation}. The greater the value of $\Delta$, the more significant the contrast is. Figure 1 displays duration histograms, associated gamma curves and notations, for phoneme /a/. 

\begin{equation}
\Delta = \argmax{x}(d_L(x)) - \argmax{x}(d_S(x))
\label{delta_equation}
\end{equation}


Finally, it is important to note that we did not use Hartigan's Dip test of unimodality \cite{hartigan1985dip} since our preliminary measurements have shown that this test always concludes to the bi-modality of our distribution - even for extremely weak contrasts.

\section{Machine Assisted Analysis of Vowel Length Contrasts in Wolof}
\label{experiments}

	\subsection{Data and ASR System}


In addition to our existing in-house (Dakar standard
) Wolof read speech corpus \cite{gauthier2016collect}, we 
recently collected data during a field trip in Senegal.
We collected semi-spontaneous speech of Wolof (Dakar standard) and dialectal variants. In total, we gathered around 1.5 hours of elicitated speech from 22 speakers (6 Faana-Faana speakers, 2 Lebu speakers, 3 speakers of urban Wolof and 11 speakers of standard Wolof). Each speaker had to watch a series of 76 short videos designed to express trajectory \cite{grinevald2011constructing}. This data can be considered as semi-spontaneous speech.

Our best Wolof ASR system was used to decode new recorded speech. This is a standard context dependent DNN-HMM hybrid system trained with Kaldi speech recognition toolkit \cite{povey2011kaldi}. More details on this system can be found in \cite{interspeech-gauthier2016} and it is made available through a VM\footnote{see \url{https://github.com/besacier/ALFFA_PUBLIC/blob/master/ASR/WOLOF/WOLOF-VM/}}.
We used 5 transcriptions of Faana-Faana (over 6) and 3 transcriptions of standard Wolof (over 11), because only a subset of ASR hypotheses were corrected by Wolof linguists.
Table \ref{wolof_speech_corpus} summarizes each data set on which we will measure vowel length contrast in this paper.

\begin{table}[h]
\centering
\caption{{\it Wolof speech data overview.}}
\begin{scriptsize}
\renewcommand{\arraystretch}{1.2}
	\begin{tabularx}\linewidth{ p{2cm} | c | c | c | c | c }
	\bf Data Set & \bf Male & \bf Female & \bf \#Utt & \bf \#Words & \bf Duration \\ \hline
	Wolof (read) & 8 & 6 & 1,120 & 10,461 & 1h12 mins \\ \hline
	Wolof (semi-spontaneous) & 2 & 1 & 254 & 2,825 & 14 mins \\ \hline
	Faana-Faana (semi-spontaneous) & 5 & 0 & 454 & 3,365 & 19 mins \\
	\end{tabularx}
\end{scriptsize}
\label{wolof_speech_corpus}
\end{table}



	\subsection{Analysis on Wolof Read Speech}


		\subsubsection{Forced Alignment with Human Transcriptions}
\vspace{-0.1cm}
		
\begin{figure}[h!]
	\centering
	\includegraphics[scale=0.3, trim=0mm 0mm 0mm 0mm, clip=true]{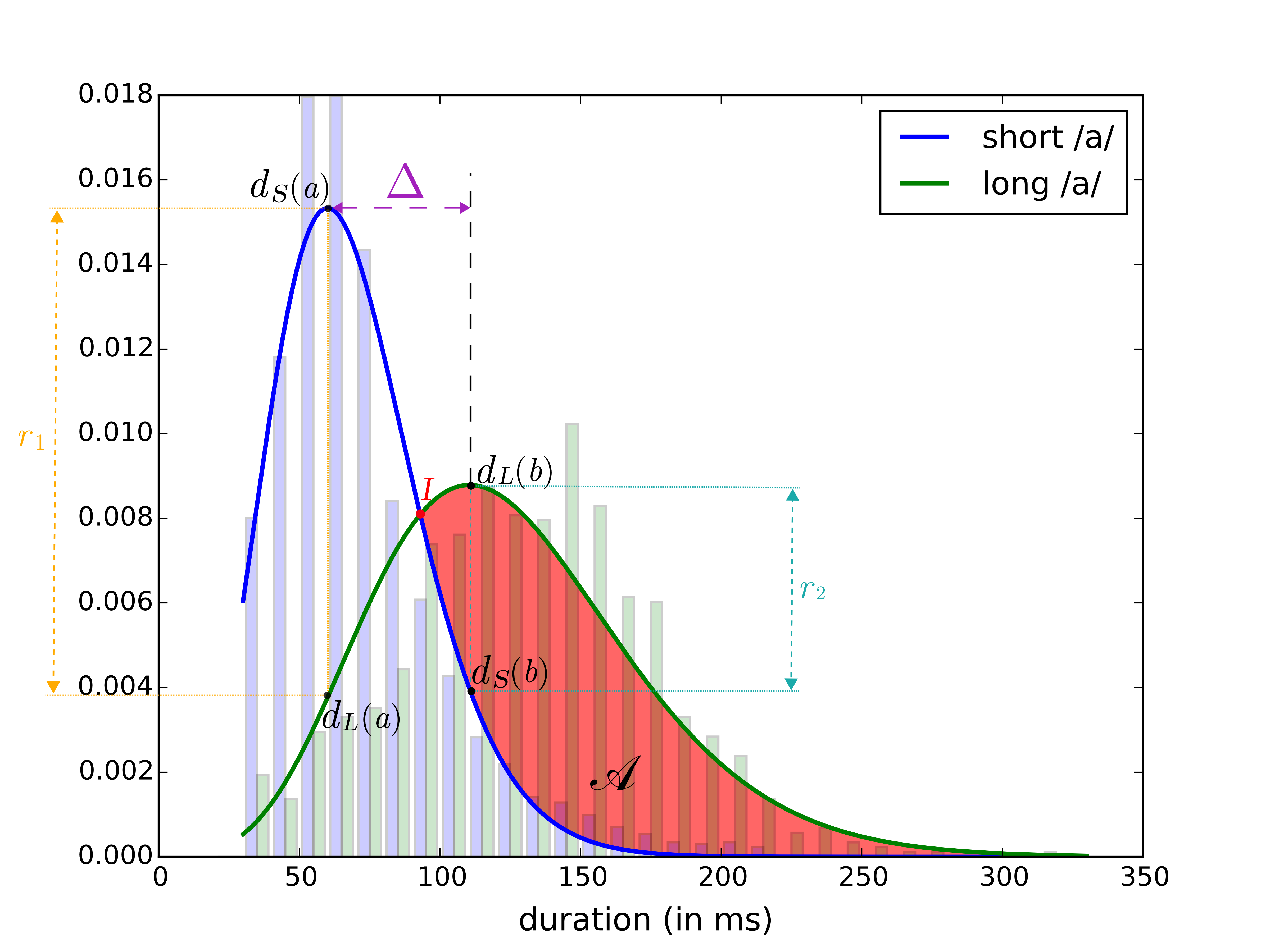}
	\caption{{\it Histogram and Gamma Distribution for vowel /a/ in Wolof Read Speech - Strong Contrast}}
	\label{strong_degree_of_bimodality}
\end{figure}

\begin{figure}[h!]
	\centering
	\includegraphics[scale=0.3, trim=10mm 2mm 10mm 10mm, clip=true]{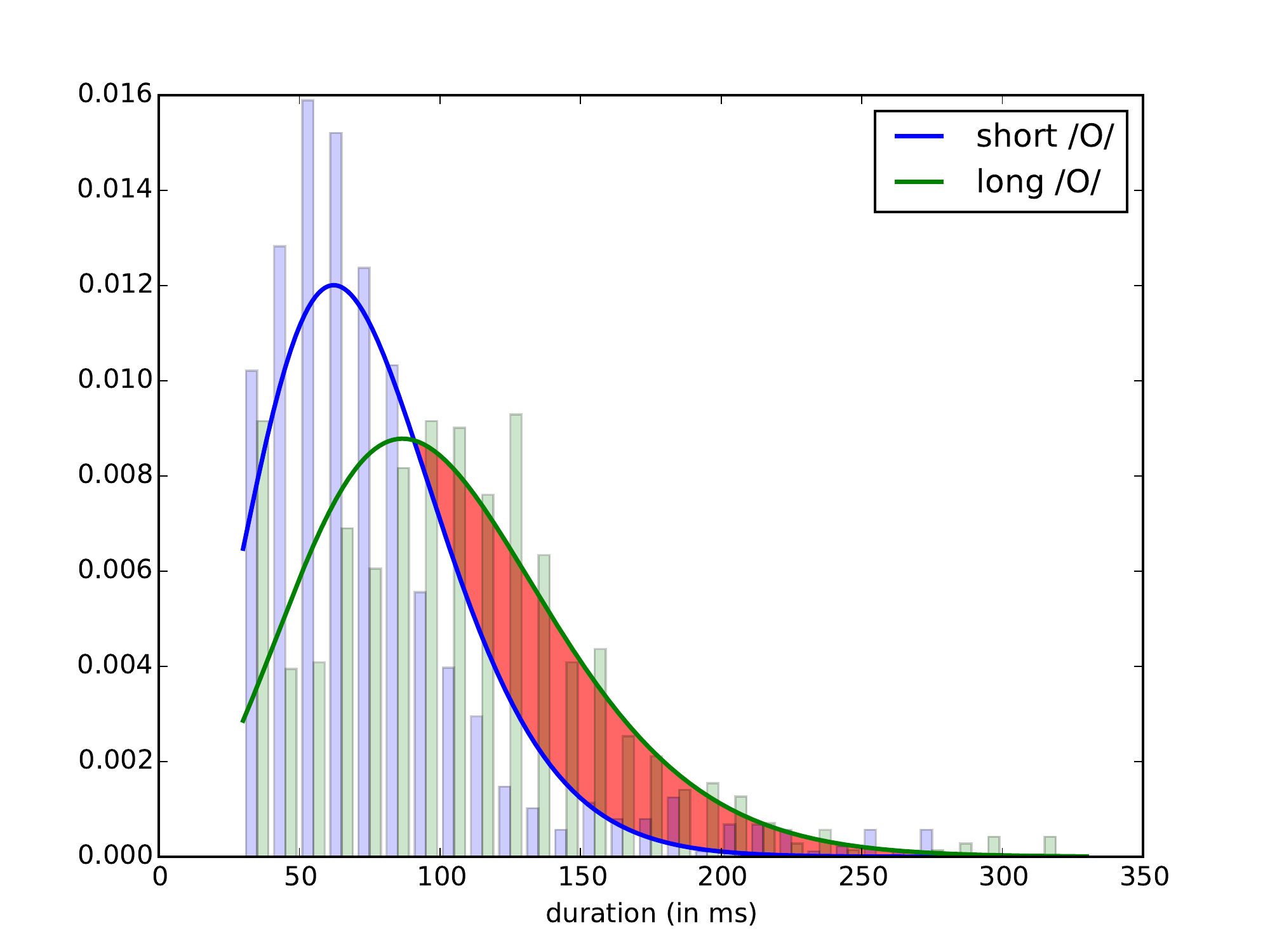}
	\caption{{\it Histogram and Gamma Distribution for vowel \textipa{/O/}  in Wolof Read Speech - Weak Contrast}}
	\label{low_degree_of_bimodality}
\end{figure}

		In a first phase, we extract vowel durations by force-aligning human transcriptions of  development (\textit{dev}) set described in \cite{interspeech-gauthier2016} (1,120 utterances, 1h12mn of speech) and made up of Wolof read speech (see Table \ref{wolof_speech_corpus}).
		Forced-alignment is done with our CD-DNN-HMM-based acoustic model (length contrasted acoustic models with different units for short and long vowels). The 7 contrasted vowels are tagged as /short/ or /long/ depending on the duplication of the grapheme within the word. 
%
Data is partitioned in different sets denoted by $ \mathscr{D}^{v}_{l} $ where $ v \in \mathcal{V} = \lbrace i,e,\text{\textipa{E}},a, \text{\textipa{O}},o,u\rbrace $ is the studied vowel and $ l \in \mathcal{L} = \lbrace S,L \rbrace $ is the expected length of the vowel (short or long). We computed vowel durations and built their histogram for each vowel after deleting outliers (we keep observations $x$ so that $ \mu-3\sigma < x < \mu+3\sigma $). We also approximate our real distribution by the probability density function of a \textit{Gamma} distribution. 
Eye-looking at normalized distributions for each vowel confirms that bimodality exists for all of them. However, the degree of contrast differs for each vowel. For instance, strong duration contrast is observed for vowel /a/ (Figure \ref{strong_degree_of_bimodality}) whereas weak contrast is observed for vowel \textipa{/O/} (Figure \ref{low_degree_of_bimodality}).

\vspace{-0.3cm}

\vspace{-0.5cm}

\vspace{-0.3cm}

\begin{table}[h!]
\centering
\caption{{\it Contrast Features Extracted on Wolof Read Speech.}}
\renewcommand{\arraystretch}{1.2}
	\begin{scriptsize}
		\begin{tabular}{c|c|c|c|c|c|c}
		\label{table_dev_ref}

\textbf{Phoneme} & \centering \multirow{3}{*}{\bf \#occurences} & \multirow{3}{*}{$\mathbf{\mu}$} & \multirow{3}{*}{$\mathbf{r_1}$} & \multirow{3}{*}{$\mathbf{r_2}$} & \multirow{3}{*}{$\mathbf{\mathscr{A}}$} & \multirow{3}{*}{$\Delta$} \\ 
short & & & & & &  \\
long & & \tiny{(in ms)} & & & & \tiny{(in ms)} \\ \hline

\bf \textipa{/i/} & 2,149 & 76 & \multirow{2}{*}{ \bf 2.54} & 
\multirow{2}{*}{ \bf 1.42} & 
\multirow{2}{*}{ \bf 0.44} & 
\multirow{2}{*}{ \bf 49} \\
\bf \textipa{/i:/} & 133 & 131 & & & & \\ \hline

\bf /e/ & 227 & 79 & \multirow{2}{*}{ \bf 2.63} & 
\multirow{2}{*}{ \bf 1.52} & 
\multirow{2}{*}{ \bf 0.45} & 
\multirow{2}{*}{ \bf 37} \\
\bf \textipa{/e:/} & 178 & 120 & & & & \\ \hline

\bf \textipa{/E/} & 1,264 & 81 & \multirow{2}{*}{ \bf 2.64} & 
\multirow{2}{*}{ \bf 1.50} & 
\multirow{2}{*}{ \bf 0.45} & 
\multirow{2}{*}{ \bf 46} \\
\bf \textipa{/E:/} & 557 & 131 & & & & \\ \hline

\bf \textipa{/a/} & 4,673 & 69 & \multirow{2}{*}{ \bf 4.07} & 
\multirow{2}{*}{ \bf 2.21} & 
\multirow{2}{*}{ \bf 0.56} & 
\multirow{2}{*}{ \bf 50} \\
\bf \textipa{/a:/} & 880 & 125 & & & & \\ \hline

\bf \textipa{/O/} & 881 & 73 & \multirow{2}{*}{ \bf 1.62} & 
\multirow{2}{*}{ \bf 0.93} & 
\multirow{2}{*}{ \bf 0.27} & 
\multirow{2}{*}{ \bf 24} \\
\bf \textipa{/O:/} & 710 & 102 & & & & \\ \hline
 
\bf \textipa{/o/} & 60 & 68 & \multirow{2}{*}{ \bf 2.85} & 
\multirow{2}{*}{ \bf 1.27} & 
\multirow{2}{*}{ \bf 0.46} & 
\multirow{2}{*}{ \bf 34} \\
\bf \textipa{/o:/}  & 69 & 108 & & & & \\ \hline
 
\bf \textipa{/u/} & 1,893 & 67 & \multirow{2}{*}{ \bf 2.34} & 
\multirow{2}{*}{ \bf 1.09} & 
\multirow{2}{*}{ \bf 0.40} & 
\multirow{2}{*}{ \bf 36} \\
\bf \textipa{/u:/} & 111 & 110 & & & & \\

		\end{tabular}
	\end{scriptsize}
\end{table}


Table \ref{table_dev_ref} shows measurements of length contrast. Vowels are sorted according to their height. In addition to the contrast features described in Section \ref{features}, we also display in third column the mean duration $\mathbf{\mu}$ (in ms) for each short and long vowel. Vowel /a/ is the one that appears most frequently (both short and long) while vowel /o/ is the one that appears most rarely. 
This is easily explained because words containing the vowel /a/ are very common while those containing vowel /o/ are rare in Wolof.
We observe that 2 articulatory features affect vowel duration: height and backness. Indeed, mean duration of short vowels increases with the aperture of the jaw, as described in \cite{sock1983organisation}, except for /a/. The phonological status of /a/ is still in debate and \cite{cisse2006problemes} raises the fact that linguists are not all unanimous on the issue. The same rule is not observed on long vowels. Mean duration also shows that back vowels (\textipa{/O/}, /o/ and /u/) are shorter than front vowels (/i/, /e/, \textipa{/E/}), for both short and long phonemes.\\
$\Delta$ varies from 24 ms to 50 ms and $\mathscr{A}$ from 0.27 to 0.56. Vowel /a/ is the one with the strongest length contrast, with large $r_1$ and $r_2$ ratios, as well as large area $\mathscr{A}$ and large $\Delta$. Though \textipa{/O/} is the vowel with the least distinguishable length contrast, with low $r_1$ and $r_2$ ratios, small $\mathscr{A}$ and moderate $\Delta$, features unveil that all vowels are length-contrasted. The table also shows that contrast features are correlated but they are complementary to describe the shape of the vowel length distributions.
To conclude on this sub-section, this analysis (made fully automatically on 14k vowel tokens) show that our proposed features can highlight different degrees of contrast for each vowel considered and confirm - at a larger scale - previous analyses made.
	
	\vspace{-0.20cm}
		\subsubsection{Forced Alignment with Automatic (ASR) Transcriptions}
		\label{less_supervision}
			\vspace{-0.18cm}

\begin{figure}[h!]
	\centering
	\includegraphics[scale=0.3, trim=10mm 2mm 10mm 5mm, clip=true]{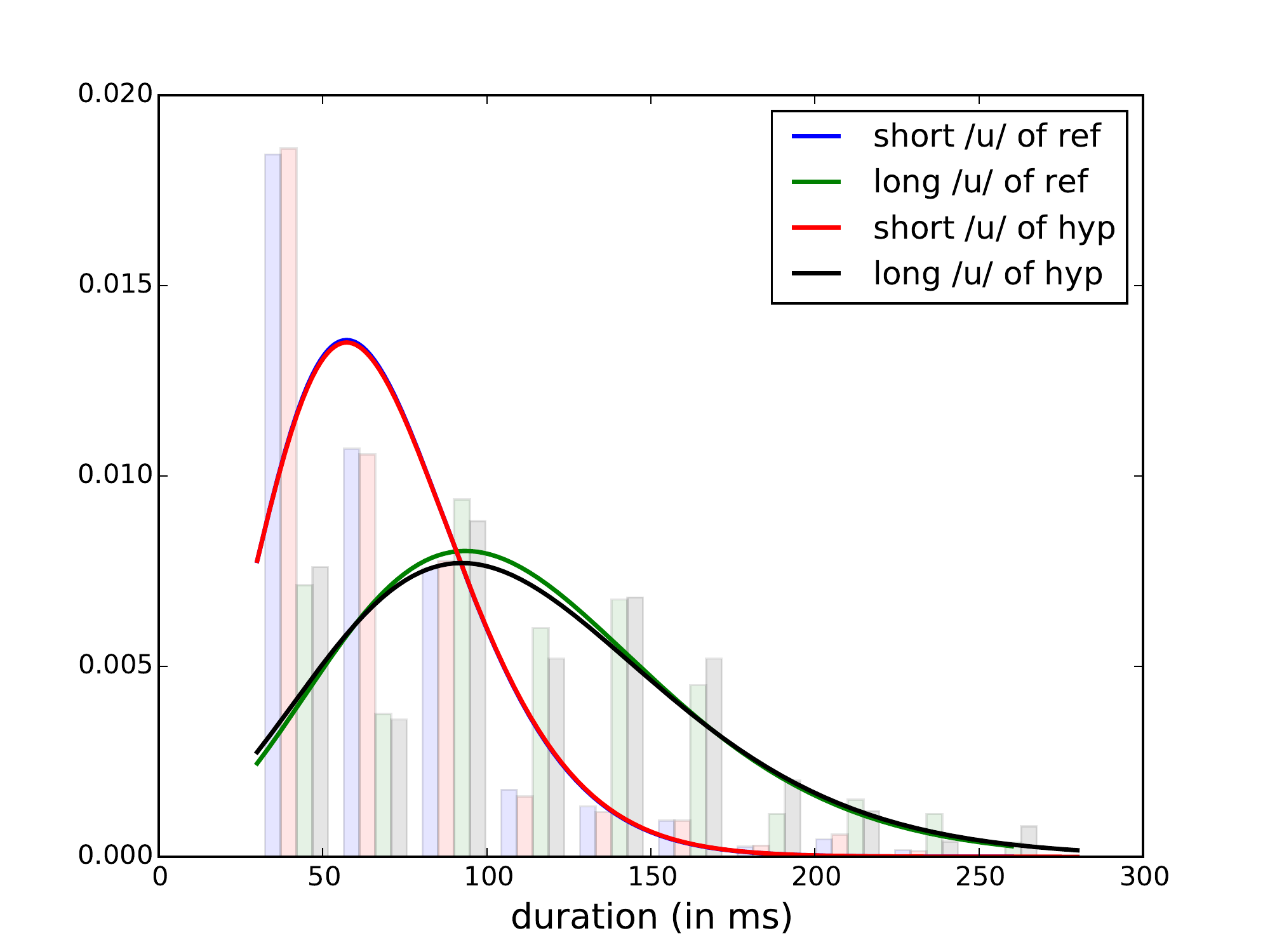}
	\caption{{\it  Histogram and Gamma Distribution for /u/ in Wolof Read Speech - Using Human (ref) or ASR (hyp) Transcripts}}
	\label{ref_vs_hyp}
\end{figure}

		In this sub-section, we try to see if manual transcriptions can be replaced by ASR hypotheses while keeping same trends/conclusions. In that case, we relax the constraint of having manual transcription of the data set. We computed vowel durations from forced alignment obtained with ASR transcripts (from our baseline Wolof ASR system, trained on held-out data - around 20\% WER on read speech) and built gamma distributions as in previous section. 
		For each vowel, we compared both distributions (manual transcription \textit{vs} ASR transcription) using Kolmogorov-Smirnov statistical test \cite{massey1951kolmogorov} (the null hypothesis $H_0$ was that both distributions obtained after manual and ASR transcriptions are similar).
	For each vowel $v$, no significant difference was found. 
To illustrate this result, Figure \ref{ref_vs_hyp} shows duration histograms and associated gamma curves for phoneme /u/ when human ($ref$) or ASR ($hyp$) transcriptions are used for forced-alignment. Both curves are very similar and this confirms that, for read speech, the need of manual transcriptions can be relaxed since the use of ASR leads to very similar measurements and to the same conclusions. For the next sub-sections (semi-spontaneous speech), ASR will be also used to produce transcripts but they will be further corrected by humans due to the more spontaneous nature of the data\footnote{Preliminary measurements have shown that the ASR transcriptions on spontaneous speech are too noisy to be used directly. We got around 31\% WER for Wolof and 66\% WER for Faana-Faana.}.
		
	\subsection{Analysis on Wolof Semi-Spontaneous Speech}
\vspace{-0.2 cm}

\begin{table}[h!]
\centering
\caption{{\it Contrast Features Extracted on Wolof Semi-Spontaneous Speech.}}
\renewcommand{\arraystretch}{1.2}
\label{table_WW-semi-spontaneous_ref}
\begin{threeparttable}
	\begin{scriptsize}
		\begin{tabular}{c|c|c|c|c|c|c}
\textbf{Phoneme} & \multirow{3}{*}{\bf \#occurences} & \multirow{3}{*}{$\mathbf{\mu}$ \tiny{(in ms)}} & \multirow{3}{*}{$\mathbf{r_1}$} & \multirow{3}{*}{$\mathbf{r_2}$} & \multirow{3}{*}{$\mathbf{\mathscr{A}}$} & \multirow{3}{*}{$\Delta$ \tiny{(in ms)}} \\ 
short & & & & & \\
long & & & & & \\ \hline

\bf \textipa{/i/} & 1,757 & 72 & \bf \multirow{2}{*}{1.06} & \bf \multirow{2}{*}{1.01} & \bf \multirow{2}{*}{0.10} & \bf \multirow{2}{*}{11} \\ 
\bf \textipa{/i:/} & 252 & 83 & & & & \\\hline

\bf \textipa{/e/} & 161 & 71 & \bf \multirow{2}{*}{1.10} & \bf \multirow{2}{*}{1.14} & \bf \multirow{2}{*}{0.19} & \bf \multirow{2}{*}{12} \\ 
\bf \textipa{/e:/} & 213 & 83 & & & & \\\hline

\bf \textipa{/E/} & 518 & 69 & \bf \multirow{2}{*}{1.40} & \bf \multirow{2}{*}{0.91} & \bf \multirow{2}{*}{0.21} & \bf \multirow{2}{*}{18} \\ 
\bf \textipa{/E:/} & 225 & 90 & & & & \\\hline

\bf \textipa{/a/} & 1,815 & 60 & \bf \multirow{2}{*}{2.56} & \bf \multirow{2}{*}{1.32} & \bf \multirow{2}{*}{0.44} & \bf \multirow{2}{*}{35} \\ 
\bf \textipa{/a:/} & 324 & 100 & & & \\\hline

\bf \textipa{/O/} & 360 & 67 & \bf \multirow{2}{*}{1.22} & \bf \multirow{2}{*}{0.84} & \bf \multirow{2}{*}{0.09} & \bf \multirow{2}{*}{5} \\ 
\bf \textipa{/O:/} & 190 & 74 & & & & \\\hline

\bf \textipa{/o/} & 62 & 51 & \bf \multirow{2}{*}{6.12} & \bf \multirow{2}{*}{3.26} & \bf \multirow{2}{*}{0.61} & \bf \multirow{2}{*}{35} \\ 
\bf \textipa{/o:/} & 123 & 89 & & & & \\\hline

\bf \textipa{/u/} & 755 & 51 & \bf \multirow{2}{*}{\tnote{\textdagger}} & \bf \multirow{2}{*}{5.95} & \bf \multirow{2}{*}{0.73} & \bf \multirow{2}{*}{44} \\ 
\bf \textipa{/u:/} & 16 & 96 & & & & \\
		\end{tabular}
	\end{scriptsize}
	\begin{tablenotes}
    \item[\textdagger] \scriptsize{The ratio can not be computed because there were no data for the long unit of the phone ($d_L(a)=0$) at point corresponding to the mode of the short phone distribution.}
	\end{tablenotes}
\end{threeparttable}
\end{table}

\vspace{-0.2 cm}
We computed same features shown in Table \ref{table_dev_ref} on our Wolof semi-spontaneous corpus. Results are presented in Table \ref{table_WW-semi-spontaneous_ref}. 
\vspace{-0.02 cm}

Looking at the mean duration of the vowels $\mathbf{\mu}$, our first remark is that it is lower in semi-spontaneous speech compared to read speech (for both short and long units). These conclusions were expected but they confirm that our machine-assisted methodology allows usable measurements at a larger scale.
Comparing $\mathbf{\mu}$ in read and semi-spontaneous context, we observe that long vowels are the most affected by the speaking style, especially front vowels (\textipa{/i:/}, \textipa{/e:/} and \textipa{/E:/}), while short units are the least impacted among the vowel set.
Results for /u/ have to be taken with caution, since we only have 16 long occurences, as well as for \textipa{/o/}\texttildelow \textipa{/o:/} for which we have less occurences compared to other vowels. All computed features show that length contrast on \textipa{/O/}\texttildelow \textipa{/O:/} pair is significantly reduced in semi-spontaneous speech in comparison to what was  observed in read speech.
In addition, the vowel height has no longer influence on the duration. Theses findings are consistent with \cite{gendrot2005impact} who described that spontaneous speech have an effect on the vowel pronunciation which tends to be more centralized when pronounced shorter.




\vspace{-0.2cm}
	\subsection{Analysis on a Dialectal Variant of Wolof}
	
	\vspace{-0.1cm}


\begin{table}[h!]
\caption{{\it Contrast Features Extracted on Faana-Faana Semi-Spontaneous Speech.}}
\label{table_FF-semi-spontaneous_ref}
\renewcommand{\arraystretch}{1.2}
\centering
\begin{threeparttable}
	\begin{scriptsize}
		\begin{tabular}{c|c|c|c|c|c|c}
\textbf{Phoneme} & \multirow{3}{*}{\bf \#occurences} & \multirow{3}{*}{$\mathbf{\mu}$ \tiny{(in ms)}} & \multirow{3}{*}{$\mathbf{r_1}$} & \multirow{3}{*}{$\mathbf{r_2}$} & \multirow{3}{*}{$\mathbf{\mathscr{A}}$} & \multirow{3}{*}{$\Delta$ \tiny{(in ms)}} \\ 
short & & & & & \\
long & & & & & \\ \hline

\bf \textipa{/i/} & 882 & 69 & \bf \multirow{2}{*}{0.91} & \bf \multirow{2}{*}{1.14} & \bf \multirow{2}{*}{0.09} & \bf \multirow{2}{*}{8} \\ 
\bf \textipa{/i:/} & 167 & 75 & & & & \\\hline

\bf \textipa{/e/} & 77 & 74 & \bf \multirow{2}{*}{0.87} & \bf \multirow{2}{*}{1.41} & \bf \multirow{2}{*}{0.21} & \bf \multirow{2}{*}{11} \\ 
\bf \textipa{/e:/} & 116 & 83 & & & & \\\hline

\bf \textipa{/E/} & 197 & 69 & \bf \multirow{2}{*}{1.17} & \bf \multirow{2}{*}{1.06} & \bf \multirow{2}{*}{0.18} & \bf \multirow{2}{*}{17} \\ 
\bf \textipa{/E:/} & 176 & 87 & & & & \\\hline

\bf \textipa{/a/} & 909 & 63 & \bf \multirow{2}{*}{1.76} & \bf \multirow{2}{*}{1.02} & \bf \multirow{2}{*}{0.32} & \bf \multirow{2}{*}{27} \\ 
\bf \textipa{/a:/} & 188 & 94 & & & & \\\hline

\bf \textipa{/O/} & 197 & 63 & \bf \multirow{2}{*}{1.12} & \bf \multirow{2}{*}{0.90} & \bf \multirow{2}{*}{0.06} & \bf \multirow{2}{*}{3} \\ 
\bf \textipa{/O:/} & 112 & 68 & & & & \\\hline

\bf \textipa{/o/} & 24 & 53 & \bf \multirow{2}{*}{2.76} & \bf \multirow{2}{*}{1.40} & \bf \multirow{2}{*}{0.46} & \bf \multirow{2}{*}{21} \\ 
\bf \textipa{/o:/} & 50 & 77 & & & & \\
		\end{tabular}
	\end{scriptsize}
	\begin{tablenotes}
    \item[\textdagger] \scriptsize{/u/ is not represented because we do not have enough data for a comparison.}
	\end{tablenotes}
	\vspace{2mm}
\end{threeparttable}
\end{table}

We computed same features shown in Table \ref{table_dev_ref} and Table \ref{table_WW-semi-spontaneous_ref} on our Faana-Faana semi-spontaneous corpus (see Table \ref{table_FF-semi-spontaneous_ref}).

As we can see in Table \ref{table_FF-semi-spontaneous_ref}, long vowels \textipa{/e:/} and \textipa{/o:/} still appear more frequently than their short counterpart, as in semi-spontaneous (standard) Wolof. 
We observe that the duration increases with vowel height, for front long vowels (\textipa{/i:/}, \textipa{/e:/}, \textipa{/E:/}) but not for their short counterparts.
By looking at the value of the features, we note that 
 distinction between short and long pronunciation of vowels is tenuous. 
The length contrast on vowel \textipa{/O/} is also weakened, as in semi-spontaneous (standard) Wolof.
These results do not allow to demonstrate that there exists in Faana-Faana a strong opposition of vowels length as observed in (standard) Wolof. In the mean time, we can not affirm that vowel length contrast does not exist in Faana-Faana. In the descriptions of this dialect, as in the Gambian Wolof, the short/long opposition is described, so we can hypothesize that dialectal differences in Wolof are not based on this lack of contrast.  
In addition, two-sample Kolmogorov-Smirnov tests revealed that /e/, \textipa{/E/}, /a/, \textipa{/O/} vowel distributions in semi-spontaneous Wolof data set were not found significantly different from those in semi-spontaneous Faana-Faana data set but /i/, /o/ and /u/ vowel distributions were.
 Finally, since this variant has been little studied, we hope that our analysis represent one first stone in the study of phonemic contrast in Wolof dialects.
\vspace{-0.3cm}
\section{Conclusion}
\vspace{-0.1cm}

We presented in this study a large scale analysis (compared to previous phonetic studies) of vowel length contrasts in Wolof. We worked on different speaking styles but also on one dialectal variant (Faana-Faana). We proposed correlated but complementary features to describe the shape of the vowel length distributions and to highlight different degrees of length contrast given a vowel. Another important result is that relaxing the constraints on the transcriptions (by using ASR transcriptions instead of manual transcriptions) is possible for read speech since it leads to very similar distributions of durations. Future work will be dedicated to leveraging computational models and machine learning for large scale speech analysis and laboratory phonetics.
Further work will analyze the relation between these distinctive features of the length contrast distribution and the functional load concept developed by \cite{ferragne:hal-00613604}.

%

 \section{Acknowledgements}

This work was realized in the framework of the French ANR project ALFFA (ANR-13-BS02-0009).



  \eightpt
\bibliographystyle{IEEEtran}
\bibliography{handling_or_not_duration_in_ASR}

\end{document}